\documentclass[conference,a4paper]{IEEEtran}
\IEEEoverridecommandlockouts
\usepackage{cite}
\usepackage{amsmath,amssymb,amsfonts, bm}
\usepackage{graphicx}
\usepackage{textcomp}
\usepackage{xcolor}
\usepackage[noend]{algpseudocode}
\usepackage[linesnumbered,lined,boxed,commentsnumbered,ruled]{algorithm2e}

\usepackage[utf8]{inputenc}
\usepackage[T1]{fontenc}
\usepackage{amsmath}
\usepackage{amsfonts}
\usepackage{amssymb}
\usepackage{graphicx}
\usepackage{subfigure}
\usepackage{tikz}
\usetikzlibrary{spy}
\usetikzlibrary{arrows}
\usetikzlibrary{calc,through,backgrounds,shapes}
\usepackage{pgfplots}\pgfplotsset{compat=1.9}
\usepackage{multirow}

\def\BibTeX{{\rm B\kern-.05em{\sc i\kern-.025em b}\kern-.08em
    T\kern-.1667em\lower.7ex\hbox{E}\kern-.125emX}}
\begin{document}

\title{Packet Routing with Graph Attention Multi-agent Reinforcement Learning
}


\author{\IEEEauthorblockN{Xuan Mai\IEEEauthorrefmark{1},
Quanzhi Fu\IEEEauthorrefmark{1},
Yi Chen\IEEEauthorrefmark{1}\IEEEauthorrefmark{3}
}

\IEEEauthorblockA{\IEEEauthorrefmark{1}The Chinese University of Hong Kong, Shenzhen}
\IEEEauthorblockA{\IEEEauthorrefmark{2}Shenzhen Research Institue of
  Big Data, China}
\small{E-mails: \{xuanmai,quanzhifu\}@link.cuhk.edu.cn, yichen@cuhk.edu.cn}
}
\maketitle

\begin{abstract}
Packet routing is a fundamental problem in communication networks that decides how the packets are directed from their source nodes to their destination nodes through some intermediate nodes. With the increasing complexity of network topology and highly dynamic traffic demand, conventional model-based and rule-based routing schemes show significant limitations, due to the simplified and unrealistic model assumptions, and lack of flexibility and adaption. Adding intelligence to the network control is becoming a trend and the key to achieving high-efficiency network operation. In this paper, we develop a model-free and data-driven routing strategy by leveraging reinforcement learning (RL), where routers interact with the network and learn from the experience to make some good routing configurations for the future. Considering the graph nature of the network topology, we design a multi-agent RL framework in combination with Graph Neural Network (GNN), tailored to the routing problem. Three deployment paradigms, centralized, federated, and cooperated learning, are explored respectively. Simulation results demonstrate that our algorithm outperforms some existing benchmark algorithms in terms of packet transmission delay and affordable load.
\end{abstract}


\section{Introduction}
Packet routing is a key function in communication networks which consist of a set of nodes and links that connect them. Each node decides the next-hop of the packets in its buffer, with the goal to minimize the end-to-end (E2E) delay, which is defined as the time taken for a packet to be transmitted from its birth node to the destination node. The E2E delay is mainly affected by two factors, the transmission delay over the links, and the queuing delay that the packet waits in the queues of the nodes before being processed. Therefore the routing decision made by each node is crucial in determining the E2E delay of the packets. Traditional routing protocols are mostly model-based, largely relying on the understanding of the network environment and traffic pattern. In today's increasingly complex networks, they may no longer have good performance due to the profound uncertainties and dynamic network demand.
	
	One classic traditional routing protocol is the shortest path, which always sends packets through paths that can minimize the number of hops without considering the real-time traffic condition. It does bring low latency when the traffic load is low. However, as the traffic load increases, the links shared by multiple shortest paths may suffer from terrible congestion. In particular, when the traffic demand exceeds a certain degree, the E2E delay would increase dramatically.
	
There is another routing protocol known as global centralized routing. It requires a central controller, which can observe the states of all routers in the network at every moment, and then computes an instantaneous optimal next-hop for every packet by taking the queuing delay into consideration. The routing solution can be obtained through online dynamic programming \cite{DP}. With the real-time network information, global centralized routing can achieve much better performance than the static shortest path in terms of E2E delay. However, it is only suitable for small-scale networks and hard to be applied to real-world systems, since its computational complexity grows exponentially as the scale of the networks grows. A decentralized routing protocol is more desirable that utilizes local state to strike a balance between short paths and less congested paths. 
	
 In recent years, RL and deep reinforcement learning (DRL) have made remarkable success on many complicated decision-making problems. One of the most famous applications is game-playing, e.g., Atari and AlphaGo \cite{mnih2013DRL, Mnih2015DRL, silver2017DRL}, achieving super-human performance.	The advances in RL and DRL provide a promising technique for enabling effective model-free routing control. The model-free approach does not need strong assumptions and prior knowledge about the model of the network, allowing it to tackle the uncertainties and dynamics of the network environment.

  The idea of applying RL to packet routing can be traced back to 1994 when Q-routing was first proposed \cite{boyan1994QR}. In Q-routing, each router derives its routing policy from a locally-maintained table, which is updated according to the Q-learning algorithm. As demonstrated by experiments, Q-routing significantly outperforms the nonadaptive shortest path algorithm, especially under high traffic load. From \cite{boyan1994QR}, some variants of Q-routing were investigated, including the predictive Q-routing \cite{choi1996predictive} and the confidence-based dual reinforcement Q-routing \cite{kumar1999confidence}. They allow more exploration than Q-routing and thus achieve faster convergence speed. In \cite{hybrid}, a policy gradient method was considered in combination with Q-learning. This hybrid algorithm makes an improvement over the deterministic Q-routing algorithm as a result of applying a stochastic policy. 

 Nevertheless, a major limitation of tabular Q-learning is that it becomes inefficient when the state and action space are large. Instead of learning a state-action table, the deep Q-network (DQN) provides an alternative solution by applying deep neural network (DNN) as a function approximation to learn the action-value function. DQN is beneficial to routing in two aspects: 1) DNN can deal with large input space and thus more network information can be incorporated into the state for better routing policy; 2) DNN is trained with numerous trajectories and thus can explore sufficiently the history of network dynamics. In \cite{Mukhutdinov2018DQNrouting}, a DQN-based routing was proposed and shown to have superior performance to shortest path and Q-routing. It uses the most common DNN architecture namely the fully-connected neural network. However,  a communication network is inherently represented in the form of a graph, where the routers are vertices and the links are edges. Fully-connected architecture ignores the graph structure which can be exploited. 

In this paper, we take the network topology into consideration and propose a multi-agent DRL algorithm called Deep Graph Attention Network Routing (DGATR) to solve the routing problem. DGATR is a value-based DRL algorithm which is inspired by the recent advances in GNN and graph attention network (GAT) \cite{velikovi2017GAT}.  GNN and GAT work directly on the graph and can help to extract the relational representation between each node and its neighboring nodes.
In our design, each router in the network is treated as an independent agent and makes its own routing decisions. We introduce a novel GAT-based architecture to perform routing by just leveraging some local information at each router. The idea is that each router propagates and aggregates its local information according to the graph structure to obtain a more abstract representation of its state to make routing decisions.   

 On one hand, RL offers enormous advantages for routing. On the other 
 hand, the development of new network techniques is also providing
 fertile ground for RL deployment. For example, the development of
 software-defined network (SDN) makes it possible to implement a
 centralized intelligent routing algorithm. However, the centralized
 paradigm still stands as an open issue as the network scale expands,
 therefore, motivating us to balance between centralized
 and decentralized operations. 
 In this paper, three deployment paradigms are discussed: centralized learning, federated learning, and cooperated learning. In centralized learning, the local experience of all routers are uploaded to a central controller to train a single DGATR model. In federated learning, each router trains a local model on its own experience and uploads the model updates to a central controller which aggregates the received updates and sends back an updated global model to all routers. Cooperated learning further eases the need for a central controller, and allows the routers to exchange model updates with their neighbors.
We evaluate our DGATR under two network typologies. Experiment results show that our algorithm can tune the policy to better adapt to the dynamic
 network demand and achieve better performance compared with the
 existing RL-based algorithms in terms of E2E delay and network affordable load.

 The rest of the paper is organized as follows. Section \ref{section:formulation} introduces the RL formulation for packet routing. Section~\ref{section:DDGAN} describes the DGATR architecture. Section III gives the simulation setup and results, and Section IV concludes this paper and highlights areas of future work.

 	\begin{figure*}[ht]
		\centering
		\subfigure[Irregular $6 \times 6$ network topology]{
			\centering
			\includegraphics[width=5cm, height=4cm]{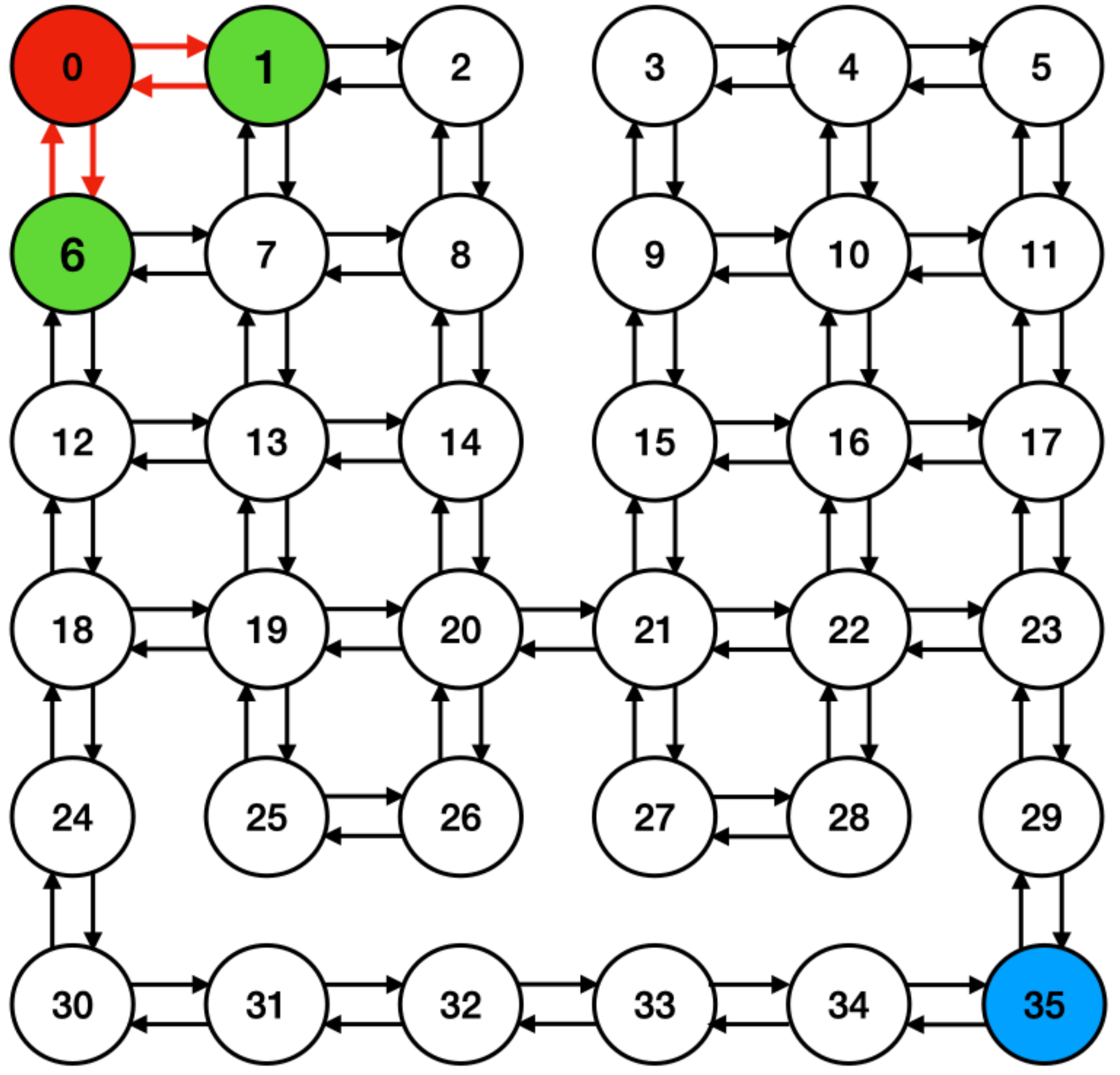}
			\label{fig:topology}}
		\hspace{1cm}
		\subfigure[DGATR neural network structure]{
			\centering
			\includegraphics[width=10cm, height=4cm]{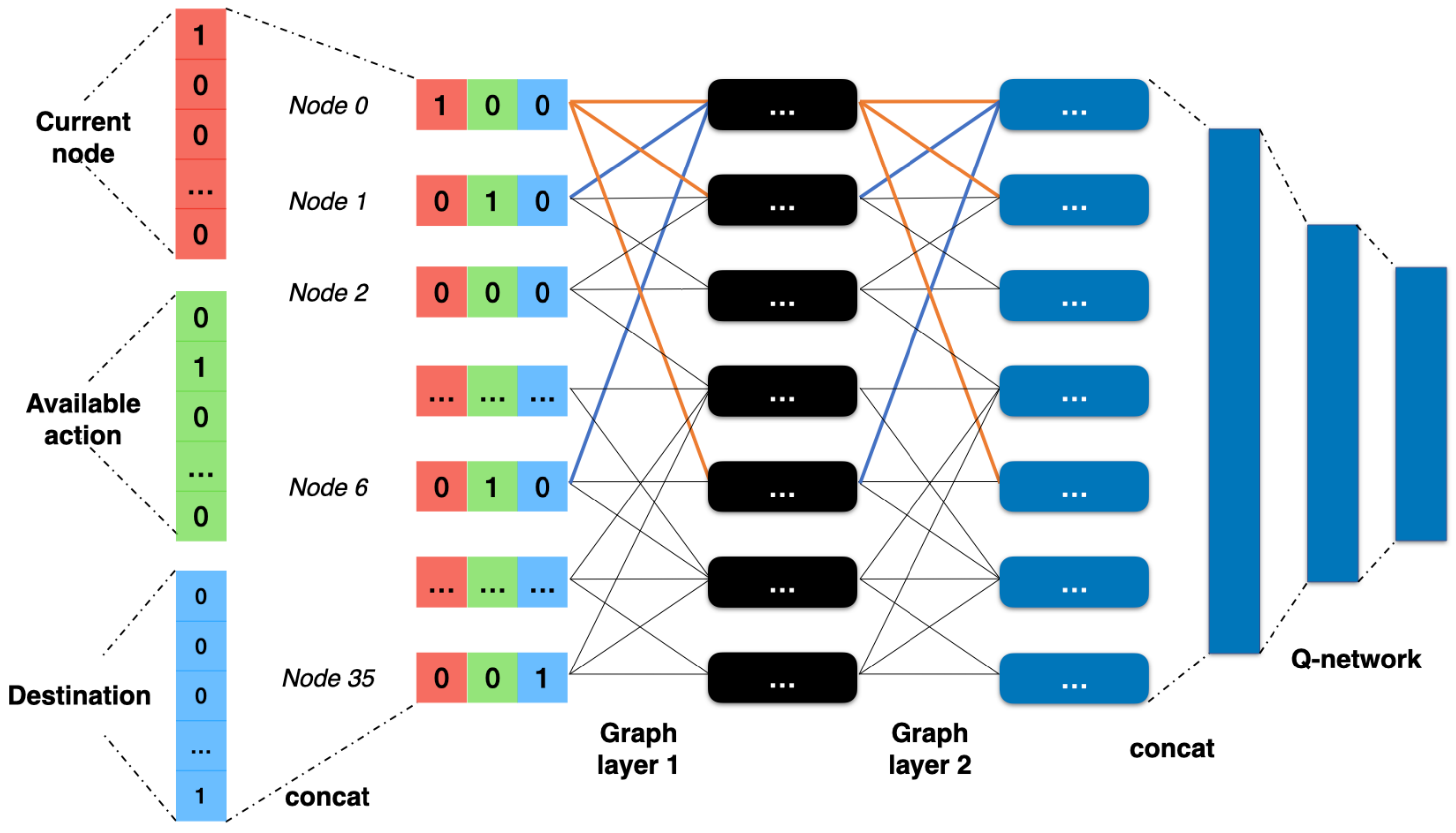}
			\label{fig:DGATR}}
		\caption{Connection between network topology and neural network structure. Example: node $0$ has a packet to node $35$ and the available next-hops are node $1$ and $6$. This local information is propagated and aggregated according to the graph. } 
	\end{figure*}

     \section{RL Formulation for Packet Routing}\label{section:formulation}
 	Consider a network consisting of a set of nodes and links which connect them. Information is transferred from one node to another node as data packets. Routing is the decision that directs data packets from their source nodes toward their destination nodes through some intermediate nodes. The goal of routing is to efficiently utilize the paths and minimize the delivery time of packets.
	Typically a network can be represented as a directed graph $\mathcal G = (\mathcal N,\mathcal E)$, where $\mathcal N = \{1,\ldots, N\}$ defines the set of nodes and $\mathcal E$ defines the set of links. An example of the network topology is shown in Fig. \ref{fig:topology}, where there are $36$ nodes (routers) and the links are bidirectional. We consider that a node can only process one packet at a time in the first-come-first-served fashion.
	
	We model the packet routing problem as a multi-agent extension of the partially observable Markov decision process (MA-POMDP).  Each node in the network is regarded as an independent agent and makes its own routing decisions through sending a packet to a possible next node. The state of the whole network is a union of all agent’s observations. Since each agent only has local observation but not global observation, the decision-making problem is based on a partially observable state. Moreover, the next state of the entire network only depends on its previous state and the actions of all the agents. 
	
	Consider the $i$th agent for $i\in \mathcal N$ and an RL framework. At each discrete time step $t$, the $i$th agent observes the local state of the environment $s_{i}^t$ and executes an action $ a_i^t$ according to its policy. Then it receives an immediate reward $r_i^t$ and the environment enters a next state $\tilde{s}_i^{t+1}$. The transition $<s_{i}^t, a_i^t, r_i^t, \tilde{s}_i^{t+1}, f_i^t>$ is explained in details as follows.
	\begin{itemize}
		\item $s_i^t$ denotes the state of agent $i$, including the available local information at the $i$th node. To be specific, the state includes the current node, the neighboring nodes of the current node, and the destination node of the first packet in the current node's queue. We use three $N$-dimensional vectors to represent this information, respectively. Each entry of the vector corresponds to one node. A value of ``1'' in the vector indicates that the node is the current node or neighboring node or destination node, and ``0'' otherwise. An example is shown in Fig. \ref{fig:DGATR}
		\item $a_i^t$ is the action of agent $i$, that is, the next-hop neighboring node that it decides to forward its packets to. 
		\item $r_i^t$ is the reward that agent $i$ receives after sending the packet to neighboring node $i'=a_i^t$. It is defined as:
		\begin{equation}\label{reward}
    	r_i^t = -(q_i^t + g_i^t),
    	\end{equation}
    	where $q_i^t$ denotes the estimated queuing delay for that packet at agent $i'$ and $g_i^t$ denotes the transmission time between two routers $i$ and $i'$.
		
		\item $\tilde{s}_i^{t+1}$ is the state of the currently delivered packet that is observed by the next-hop agent $i'$.
		\item $f_i^t$ is the transition flag indicating whether the packet arrives at its destination or not. It is defined as
		\begin{equation}\label{dest flag}
		f_i^t = 
		\begin{cases}
		1, & \text{ if $a_i^t$ is the destination}\\
		0, & \text{otherwise}
		\end{cases}.
		\end{equation}
	\end{itemize}
	
	

        The agents interact with the network environment to learn a reward-maximum policy. In this paper, cooperation among the agents is achieved either via the help of a central controller or the communication between neighboring nodes. For the former, depending on the role of the central controller, we have two deployment paradigms: centralized learning and federated learning. For the latter, we have cooperated learning. In the next section, we will first introduce the framework of the RL algorithm and then explain the three different paradigms.
        
	 \section{Deep Graph Attention Network Routing}\label{section:DDGAN}

	In this section, we introduce our DGATR algorithm, a variant of the value-based DRL routing algorithm. DGATR enables the local observations of the routers to be transformed to some more abstract and relational representations, and therefore facilitates the routing decision making. In particular, the transformation leverages the graph attention mechanism~\cite{velikovi2017GAT}. 
	
	Let us use Fig. \ref{fig:DGATR} to explain the network structure of DGATR. Taking router $i=0$ as an example, suppose it has a packet destined to router $35$. The local state of router~$0$ consists of three $N$-dimensional vectors, i.e., $(1,0,0,\ldots)$ indicating the current node, $(0,1,0,0,0,0,1,0,\ldots)$ indicating its neighboring nodes $1$ and $6$, $(0,0,\ldots,1)$ indicating the destination node $35$.  This local state is reorganized into a feature matrix $X \in \mathbb{R}^{N \times 3}$. The $j$th row of $X$, denoted by $\mathbf x_j^\intercal\in \mathbb R^3$ where $(\cdot)^\intercal$ represents transposition, now can be thought as the state of node $j$ seen by router $i$, indicating whether node $j$ is the current node, a neighboring node or a destination node, respectively. We describe a single graph attention layer but it can be easily extended to multiple layers. The input to the graph layer is a set of node features $\{\bm x_1,\ldots, \bm x_N\}$. The graph layer produces a new set of node features $\{\bm x'_1,\ldots, \bm x'_N\}, \bm x'_j\in \mathbb R^H$, where $H$ is the number of features for each node.
	
	In order to transform the input features into some more informative representations, a learnable linear transformation parametrized by weight matrix $W\in \mathbb{R}^{H \times 3}$ is first applied to all nodes:
	\begin{equation}
		\bm{h}_i = W\bm x_{i},\ i=1,2,\ldots,N.
	\end{equation}
     Let $\mathcal{N}_{i}$ be the set containing node $i$ and its neighboring nodes, e.g., $\mathcal N_0=\{0,1,6\}$ in Fig.~\ref{fig:topology}. We compute the attention coefficient of node $j \in \mathcal{N}_{i}$ to node $i$ as follows: 
	\begin{equation}\label{attention mechenism}
	\alpha_{ij} = \frac{\exp \big(\sigma(\mathbf a^\intercal [\bm{h}_i ||\bm{h}_j ])\big)} 
	{\sum_{k \in \mathcal{N}_{i}} \exp \big (\sigma(\mathbf a^\intercal [\bm{h}_i ||\bm{h}_k ]) \big)},
	\end{equation}
	where ``$||$'' is the concatenation operation; $\mathbf a\in \mathbb{R}^{2H}$ is a learnable weight vector; $\sigma(\cdot)$ is the LeackyReLu activation function.  $\alpha_{ij}$ indicates the importance of node $j$’s features to node $i$. The attention coefficients are then used to compute the final output features for every node: 
	\begin{equation}\label{gragh aggregation}
	\bm x'_{i} = \sigma(\sum_{j\in \mathcal{N}_{i}} \alpha_{ij}\bm{h}_j ), \ i=1,2,\ldots,N,
	\end{equation}
	where $\sigma(\cdot)$ is applied component-wise. It needs to be emphasized that the new features $\{\bm x'_1,\ldots, \bm x'_N\}$ are extracted solely from the local observation $X$. The graph attention network structure is tailored to achieve relational reasoning and able to abstract more meaningful information of the local observation.

\begin{algorithm}[t]
	\caption{DGATR under cooperated learning}
	\label{alg:Cooperated}
	\textbf{Input:}
	Discount factor $\gamma$, step size $\alpha$ and update weight $\tau$. \\
	\textbf{Initialize} $\mathcal D_i$ with capacity $M$, $\bm{w}_i (0)$, $\bm{w}_i^{target} (0)$ for all $i \in \mathcal N$.\\
	\textbf{Training\\}
	\For{each step $t$}{
		\For{Active Router $i \in \mathcal N$}{
			Observe the first packet and decides its next-hop based on Equation \eqref{action policy};\\
%
%
			Store the transition $(s_i,a_i , r_i,\tilde{s}_i, f_i)$ in local memory $\mathcal D_i$.\\
			
			\If{$\mathcal D_i$ is full}{
				Sample a random batch from $\mathcal D_i$ and clear the memory.\\
				Compute the loss $\mathcal{L}(\bm{w}_i (t))$ based on Equation \eqref{distributed TD} and \eqref{distributed estimated}.\\
					
					
					
				
				\textbf{Main network update}:
					$\bm{\tilde w}_i =  \bm{w}_i (t) - \alpha \nabla_{\bm{w}_i}  \mathcal{L}(\bm{w}_i (t))$.\\
				\textbf{Target network update}: $\bm{w}_i^{target} (t+1) = \tau \bm{\tilde w}_i+ (1-\tau) \bm{w}_i^{target}(t)$.
			}
		}
		\textbf{Consensus update}:\\
		$\bm w_i(t+1) = \sum_{j \in \mathcal{N}_{i}} W_{i, j} \bm {\tilde w}_j $ for all $i \in \mathcal N$
	}
\end{algorithm}
	
The extracted features are then fed into two fully connected neural networks to produce the Q-value of state-action pairs. In order to allow the neural network structure to be applied to agents with different sizes of action space (i.e., number of neighboring nodes), we fix the dimension of the output of the neural network to be $N$. Each output corresponds to one of the $N$ nodes as a possible action. For notation simplicity, all the parameters of the DGATR neural network are collectively denoted by $\bm w$. The agents share the same DGATR structure but could have different parameter values. So we use $\bm w_i$ to indicate the parameters of agent $i$'s neural network.
For agent $i$, we use $ Q_i(s_i,a;\bm w_i)$ to denote the $a$-th output of the neural network given an input $s_i$. With the chosen reward defined in \eqref{reward}, the Q-value $ Q_i(s_i,a;\bm w_i)$ indeed estimates the expected time-to-arrival (with a negative sign) that it takes to route a packet from node $i$ to the destination node specified by $s_i$ by way of a next-hop node $a$. 
Once obtained the Q-value, the agent will forward the packet to its neighboring node according to a greedy algorithm:
\begin{equation}\label{action policy}
	a_i = \arg \max_{a \in \mathcal{N}_i \setminus \{i\}}Q_i(s_i, a;\bm w_i).
	\end{equation}
	After executing this action, agent $i$ receives feedback from the next-hop agent with the reward $r_i$, its observation of the current packet $\tilde{s}_i$ and the transition flag $f_i$. As agent $i$ is interacting with the environment, a sequence of experience tuples $(s_i^t, a_i^t, r_i^t, \tilde{s}_i^{t+1}, f_i^t)$ is recorded into a memory. 
	
	Though each agent makes individual routing decisions, their joint actions together determine the E2E delay of packets. Therefore we consider three learning paradigms for efficient multi-agent RL exploration by sharing experience or model updates amongst agents. They are centralized learning, federated learning, and cooperated learning. Next, we will introduce them one by one.
	
	
	\subsection{Centralized learning}
	In the centralized paradigm, a central controller is assumed to be available. The central controller collects the experience from all agents and stores them together in a large centralized memory $\mathcal D$ to train a single DGATR model, that is, $\bm w_i = \bm w$ for all $i\in \mathcal N$. Each agent independently uses the same copy of the DGATR model to make routing decisions. 
	
	When training the DGATR, we use the techniques of experience replay and 
	target network \cite{sutton2018reinforcement}. Experience replay refers to the act of sampling a small batch of experience tuples from the memory to train a network, which can circumvent the issue of time-correlated samples. The target network, with parameters $\bm w^{\text{target}}$, is a duplicate of the main DGATR network and used to compute the target Q-value. The main network is trained by minimizing the loss: 
	\begin{equation}
		\mathcal{L}(\bm w) = \frac1n \sum_{j=1}^n\big(y_{j} - Q(s_{j}, a_{j}; \bm w)\big)^{2},
		\end{equation}
	 which is computed over a batch of $n$ tuples $(s_j, a_j, r_j, \tilde s_{j}, f_j)$ randomly sampled from $\mathcal D$.
Therein, $y_j$ is the target Q-value estimated according to the Bellman equation: 
	\begin{equation}\label{groud truth}
		y_{j} = r_{j} + \gamma \max_{a'}Q(\tilde s_{j}, a'; \bm w^{\text{target}}) \times (1 - f_{j}),
	\end{equation}
where $\gamma \in [0,1]$ is a discount factor. The parameters $\bm w$ of the main DGATR network are trained using gradient descent algorithm. It is important to mention that the target network's parameters are not trained, but are updated by slowly copying the parameters of the main network. For example, we apply the moving average to update $\bm w^{\text{target}}$ as follows \cite{DDPG}:
\begin{equation}\label{soft target}
			\bm w^{\text{target}}(t+1) = \tau \bm w(t+1)+ (1-\tau) \bm w^{\text{target}}(t),
			\end{equation}
			where iteration indexes $t$ and $t+1$ are added to emphasize the parameters' values before and after updates, and $\tau \in [0,1]$ is a weight of the update.
			
		After training, the central controller distributes the parameters of the main DGATR network to all routers for execution. Centralized learning enables efficient exploration by using experiences from all agents. However, it may incur excessive communication costs. Besides, when the traffic requests are distributed unevenly among the routers, the learning result may be more influenced by the experiences of the busy routers.

	\subsection{Federated learning}
	To ease the communication cost imposed by centralized learning, we consider the paradigm of federated learning. The general principle consists in training local DGATR networks on local experience and aggregating the parameters updates to generate a global DGATR network. 
Specifically, each agent possesses a local replay memory $\mathcal D_i$ for $i\in \mathcal N$ to store its own experience tuples $(s_i^t, a_i^t, r_i^t, \tilde{s}_i^{t+1}, f_i^t)$ and use them to train a local DGATR network with parameters $\bm w_i$. During each update of agent $i$, a mini-batch of $n$ tuples are randomly sampled from $\mathcal D_i$ and used to compute and minimise the loss:
	 \begin{equation}
	 \label{distributed TD}
	 	\mathcal{L}(\bm{w}_{i}) = \frac1n\sum_{j=1}^{n}(y_{j} - Q(s_{j}, a_{j}; \bm{w}_i)^{2} \text{ with}
	 \end{equation}
	 \begin{equation}
	 \label{distributed estimated}
	 	y_{j} = r_{j} + \gamma \max_{a'}Q(s_{j+1}, a'; \bm{w}_{a_j}^{\text{target}}) \times (1-f_{j}).
	 \end{equation}
 The target value $y_j$ is computed using the target network of the next-hop router $a_j$. In practice, there is no need for agent~$i$ to know $\bm{w}_{a_j}^{\text{target}}$. Instead, agent $i$ receives a feedback from agent $a_j$ about $\max_{a'}Q(s_{j+1}, a'; \bm{w}_{a_j}^{\text{target}})\cdot (1-f_{j})$ after sending a packet to it. The agent computes locally the gradient of the loss given in \eqref{distributed TD} with respect to the local parameters $\bm w_i$, i.e., $\nabla_{\bm w_i} \mathcal{L}(\bm w_i)$, and uploads it to a central controller where a global network is maintained.
The parameters $\bm w_G$ of the global network is then updated by
  \begin{equation}
  \label{distributed global update}
   \bm{w}_{G}(t+1) = \bm{w}_{G}(t) - \alpha \nabla_{\bm w_i} \mathcal{L}(\bm w_i(t)),
  \end{equation}
  where $\alpha>0$ is the step size. Later, the central controller sends $\bm{w}_{G}$ back to agent $i$ and the local parameters are synchronized with the global parameters, that is, $\bm{w}_i(t+1)=\bm{w}_G (t+1)$. The target network's parameters $\bm{w}_i^{\text{target}}$ are updated in the same way as that shown in \eqref{soft target}. Note that the global network is updated as soon as a local parameter update is available and the aggregation of gradients of all agents is done asynchronously.

Federated learning achieves efficient exploration by sharing the gradients amongst all the agents. Moreover, the asynchronous aggregation allows the less busy routers to have a less frequent update. 
	
	\subsection{Cooperated learning}

	Cooperated learning is similar to federated learning except that the parameter updates are exchanged only between neighboring agents without the orchestration of a central controller.
	Each agent $i$ trains a local DGATR network on its local experience, just like \eqref{distributed TD} and \eqref{distributed estimated}. During each update of agent~$i$, a local auxiliary parameter $\bm{\tilde w}_i$ is first computed as follows:
	\begin{equation}
		\bm{\tilde w}_i  = \bm{w}_i (t) - \alpha \nabla_{\bm{w}_i} \mathcal{L}(\bm{w}_i (t)),
	\end{equation}
	and shared to the neighbors of agent $i$ immediately. So each agent has a record of the latest $\bm{\tilde w}_j$ for all $j\in \mathcal N_i$. Then agent~$i$ locally aggregates the auxiliary parameters within its neighborhood according to:
	\begin{equation}
	\label{communication}
	\bm w_i (t+1) = \sum_{j \in \mathcal{N}_{i}} W_{i, j} \bm{\tilde w}_j ,
	\end{equation}
	where $W_{i,j}$ is the $i,j$-th entry of matrix $W = D^{-1}A$, and $A\in\mathbb R^{N\times N}$ is the adjacency matrix of the network graph and $D\in\mathbb R^{N\times N}$ is a diagonal matrix with $D_{i,i} = \sum_{j}A_{i,j}$ representing the degree of the nodes in the graph.

	Cooperated learning achieves efficient exploration by encouraging parameter sharing among neighbors. It can further reduce the communication cost.
	Due to the page limitation, we only provide the DGATR under cooperated learning in Algorithm~\ref{alg:Cooperated}. The three learning paradigms are compared in next section.

\section{Experiment Results}
	We test our DRL algorithms DGATR on the irregular $6 \times 6$ grid network as shown in Fig. \ref{fig:topology} and the real AT\&T North America network \cite{ATT}. Performance comparisons are made among our algorithms and the following five widely used baseline algorithms: 1) Shortest Paths, a static routing method that is optimal when the network load is low; 2) Q-routing \cite{boyan1994QR}, using tabular Q-learning ; 3) Hybrid \cite{hybrid}, a combination of tabular Q-learning and policy gradient method; 4) DQN-routing \cite{Mukhutdinov2018DQNrouting}, using fully-connected DNN under centralized training; 5) Global Routing, an instantaneous optimal routing. We use the average packet E2E delay and network sustainable load as the performance metrics for comparisons. Since global routing generates the globally optimal next-hop for the packets at each time step, it serves as the lower bound of average packet delay and the upper bound of the sustainable load. 
	
		We model the network environment as follows. The arrival process of the packets is simulated by a Poisson process. The network load specifies the average number of packets generated per time unit in the Poisson process. The source node and the destination node of the injected packets are chosen uniformly. Furthermore, once the packet arrives at its destination, it will be removed from the network immediately. The queue that multiple packets line up at routers follows First-In-First-Out (FIFO) rule. That is, the router always processes the top packet in the queue. 
For all routing algorithms, the hyperparameters are pre-tuned. Once the hyperparameters are selected, they will be fixed in all experiments. 

For DQN-routing and DGATR, we pre-train the neural networks using the off-policy technique before letting them interact with the real network environment. The transitions $(s_i^t, a_i^t, r_i^t, \tilde{s}_i^{t+1}, \tilde{a}_i^{t+1}, f_i^t)$ used to pre-train the neural networks are produced by a well-trained Q-routing policy at network load $1$. The purpose of pre-train is to accelerate the training in the initial stage. In order to see this, we try a different number of pre-train steps and evaluate the models just obtained by pre-train on the $6 \times 6$ grid network. 
Table \ref{table:offline} shows their performance in terms of average packet E2E delay. It can be seen that the models become better when receiving better pre-train, and DGATR requires less pre-train to achieve the same performance as DQN-routing.


\begin{table}[t]
  	\caption{Packet delay for models obtained by pre-train at load $1$.}
	\setlength{\tabcolsep}{0.03\linewidth}{
		\begin{tabular}{c|ccccc}
			\hline 
			\multirow{2}*{} & \multicolumn{5}{c}{Number of pre-train steps}\\
			& 1000 & 2000 & 3000 & 4000 & 5000\\
			\hline
			DQN-routing & 139.349 & 89.590 & 34.575 & 20.275  & 6.304\\
			\hline
			\textbf{DGATR} &  \textbf{55.032} & \textbf{5.968} & \textbf{5.839} &\textbf{5.746 } & \textbf{5.631}\\
			\hline
	\end{tabular}}
	\label{table:offline}
\end{table}

		
		                      
        	\begin{figure*}[t]
	        	\centering
	        	\subfigure[Performance on irregular $6\times6$ network]{
	        		\includegraphics[height=5cm,width=8cm]{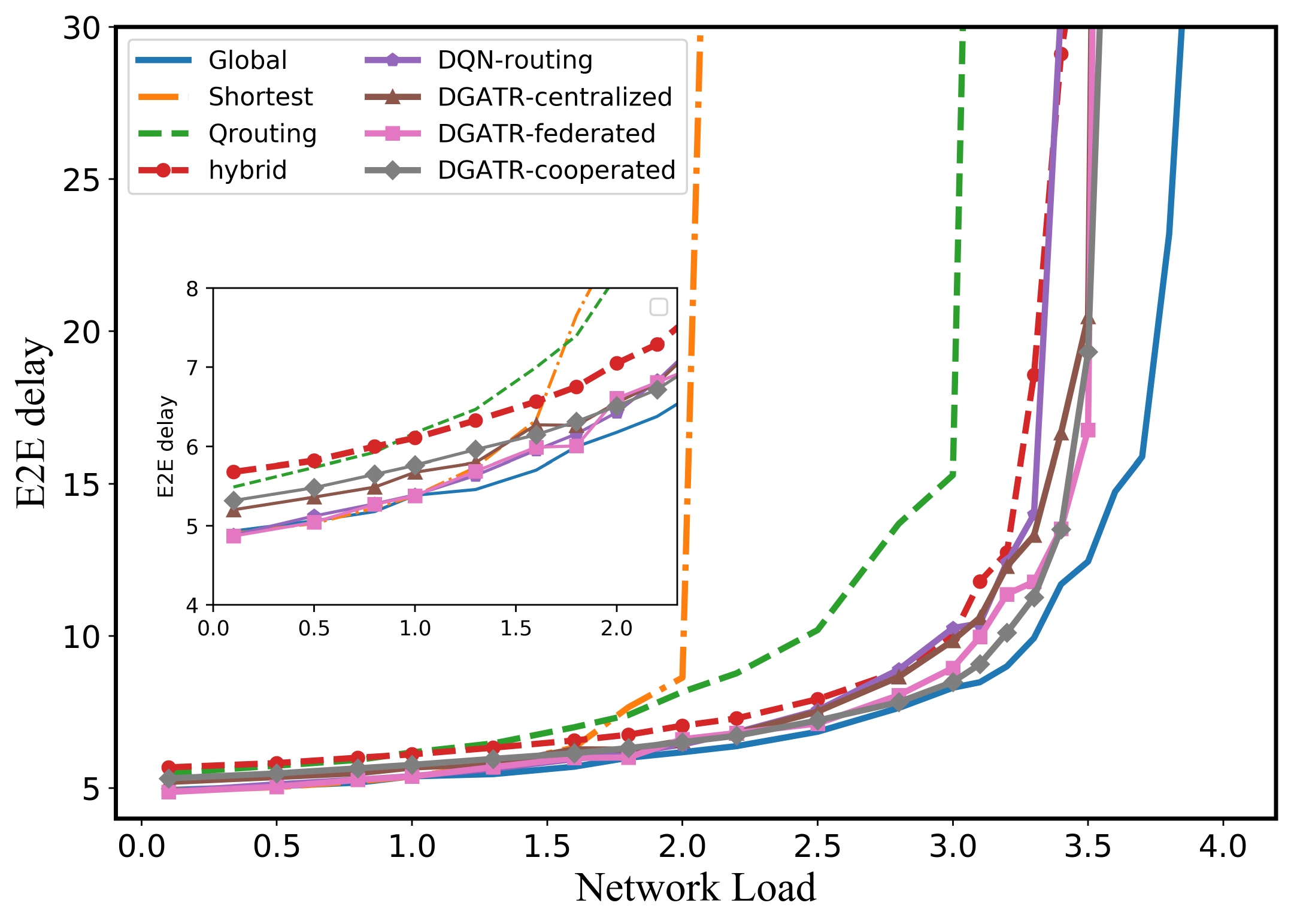}
	        		\label{fig:6x6 performance}
	        	}
	        	\hspace{0.5cm}
	        	\subfigure[Performance on real AT\&T network]{
	        		\includegraphics[height=5cm,width=8cm]{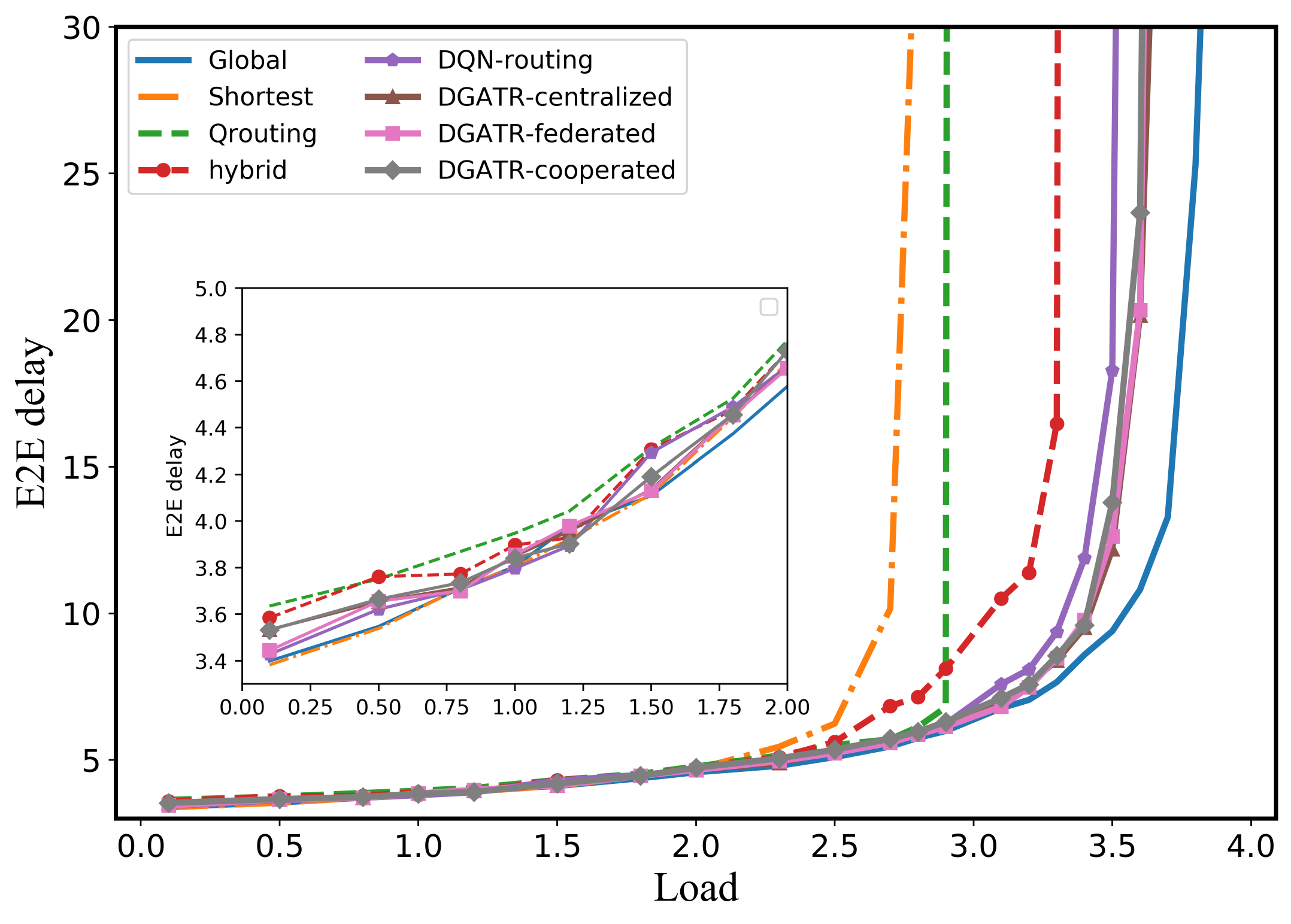}
	        		\label{fig:LATA performance}
	        	}
	        	\caption{Performance of Different Network Routing Algorithms on Two Network Topologies}
	        	\label{fig:performance}
        \end{figure*}
        \begin{table}[t]
          	\caption{Packet delay of DGATR under different learning paradigm}
                 	\centering
                 	\setlength{\tabcolsep}{0.02\linewidth}{
                 		\begin{tabular}{c|ccc}
                 			\hline 
                 			\multirow{2}*{} & \multicolumn{3}{c}{Load}\\
                 			& 3.0  & 3.2 & 3.4\\
                 			\hline
                 			DGATR-centralized & 9.845 $\pm$ 0.24  & 12.284 $\pm$ 1.14 & 16.656 $\pm$ 2.64\\
                 			\hline
                 			DGATR-distributed & 9.666 $\pm$ 0.79  & 11.4216 $\pm$ 1.33 & 13.296 $\pm$ 1.73 \\
                 			\hline
                 			\textbf{DGATR-cooperated} & \textbf{8.486 $\pm$ 0.02} &\textbf{ 10.107 $\pm$ 0.17} & \textbf{14.279 $\pm$ 1.04}\\
                 			\hline
                 			
                 	\end{tabular}}
                 	\label{tabel_load}
                 \end{table}

Fig. \ref{fig:performance} compares the average packet E2E delay of different algorithms under various network load in two network topologies.
During the experiment, we repeat varying the network load from low to high and back to low in order to simulate the network dynamics. At each load level, we record the average packet E2E delay after the learning has settled on a routing policy. The average results over ten records are reported. 
As can be seen, DGATR shows better adaptability and can sustain higher network load than the other algorithms except the Global Routing. In particular, DGATR achieves much better performance than Q-routing and Hybrid algorithms that rely on tabular Q-learning. 
The sub-figures in Fig.~\ref{fig:performance} further take a close look at the performance at low network load. Unsurprisingly Shortest Path achieves similar average packet E2E delay as the Global Routing. Moreover, the DNN-based RL algorithms are superior to the table-based RL algorithms. We understand this as following. It is widely recognized that Q-routing suffers from the lack of exploration, and therefore it fails to adapt its policy to achieve the optimal shortest paths when the network load change from high to low. In contrast, DNN as an approximation to the action-value function is trained with numerous trajectories and can explore sufficiently the history of network dynamic. In this way, DNN-based algorithms are more adaptable and resistant to the variation of the network load. Meanwhile we note that DGATR outperforms DQN-routing, suggesting that the utilization of graph information indeed can help to extract more useful features.

Last we compare the three learning paradigms for DGATR. From Fig.~\ref{fig:performance}, we can see that the maximum affordable loads of these three paradigms are indistinguishable. 
Table~\ref{tabel_load} further shows the mean and variance of packet E2E delay at three high loads on $6 \times 6$ grid network. Cooperated learning turns out to achieve lower delay and be more stable. This may because cooperated learning restricts the parameter aggregation to happen within a small neighborhood of the agents and thus the variation of the parameter update is relatively small comparing to those of centralized learning and federated learning. Moreover, all agents concurrently learn similar policies but not necessarily identical policies.

	\section{Conclusion}
	In this paper, we study the packet routing problem for communication networks. We employ DRL to devise a novel routing algorithms called DGATR. In our design, each router in the network is treated as an independent agent and makes its own routing decisions based on its local observation. GAT-based architecture is introduced to process the local observation at each router to explore the relationship of the nodes and extract latent features through the given network topology. 
	Experiment results demonstrate the superiority of our algorithm comparing to several baseline algorithms. 
	We also discuss three deployment paradigms: centralized, federated and cooperated learning, each promoting exploration at different levels.
 One of our future research directions could be to study the exploration efficiency of sharing experience and gradients for multi-agent RL. 

\bibliographystyle{IEEEtran}
	\bibliography{Routing20210421}

\end{document}